\documentclass[runningheads]{llncs}
\usepackage{geometry}
\usepackage[T1]{fontenc}
\usepackage{graphicx}
\usepackage{verbatim}
\usepackage{array}
\usepackage{amsmath}
\usepackage{amsfonts}
\usepackage{float}
\usepackage{xurl}
\usepackage{color}
\usepackage{booktabs}
\usepackage{hyperref}
\usepackage{multirow}

\usepackage{cite} 
\usepackage{multirow}
\usepackage{multicol}
\begin{document}

\title{Does Linguistic Structure Enrichment Enhance Coherence Assessment? Not With Current Architectures} 

\author{Victor Mazzotti\inst{1}\and Luiz Pereira\inst{2}\and Marina Bitencourt dos Santos\inst{3}\and Helena Maia\inst{2}\and Carlos Caetano\inst{2}\and\\Nádia Felix\inst{4}\and Sandra Avila\inst{2}}

\authorrunning{Mazzotti et al.}

\institute{Instituto de Matemática, Estatística e Computação Científica (IMECC)\and Instituto de Computação (IC)\and Instituto de Estudos da Linguagem (IEL)\\Universidade Estadual de Campinas (UNICAMP), Campinas, SP, Brasil\and Instituto de Informática (INF), Universidade Federal de Goías (UFG), Goiânia, GO, Brasil
}

\maketitle         
\begin{multicols}{2}
\begin{abstract}
Recent advances in large language models have transformed human–computer interaction. Despite their fluency, these models often produce texts that are grammatically correct but semantically incoherent, containing contradictions or disruptions in logical flow. This work investigates whether enriching text with syntactic and rhetorical information can improve incoherence prediction. Our experiments and analysis show that plain texts achieved higher accuracy because the added information was structurally and syntactically incompatible with the language model's architecture. Additionally, to demonstrate the practical importance of coherence assessment, we performed zero-shot experiments on a Brazilian disinformation dataset,  suggesting that textual coherence can serve as a proxy for detecting misleading content. Code and models are available at \href{https://github.com/ittozzamV/cohereclassifier}{github.com/ittozzamV/cohereclassifier}. 

\end{abstract}

\section{Introduction}
\label{sec:intro}

The growing adoption of language models has transformed how computational systems interact with humans~\cite{lee2023evaluating,zhang2023large}, enabling the generation of more fluent, contextually rich text. These systems have evolved from simpler models, such as BERT~\cite{devlin2019bert}, to large-scale architectures, such as the recent GPT-5~\cite{openai2025gpt5}. Driven by models with intuitive communication interfaces, such as ChatGPT~\cite{chatgpt} and Copilot~\cite{copilot}, these systems enable realistic conversational interactions and support automatic content generation across diverse knowledge domains. Despite these impressive capabilities, a critical challenge remains: ensuring that the generated texts are textually coherent~\cite{vakulenko2018measuring,zhao2023generation}.

Textual incoherence --- characterized by internal contradictions, abrupt topic shifts, or flaws in logical structure --- compromises the usefulness of these systems in sensitive applications, such as journalistic content production or education, where clarity and consistency are essential~\cite{fatima2022systematic,thompson1986readability,redeker2000coherence}. While traditional metrics for assessing grammatical correctness are well established in the literature~\cite{napoles2015ground,choshen2018automatic,park2020comparison}, evaluating coherence remains a challenge, particularly in languages such as Portuguese, where annotated resources \mbox{are scarce.}

To address this challenge, we propose to investigate an approach that combines well-esta\-blished linguistic theories, such as Rhetorical Structure Theory~(RST)~\cite{mann1988rhetorical}, which analyzes rhetorical relations between text segments, and grammatical classes commonly referred to as Part-Of-Speech (POS)~\cite{pradhan2005semantic}, 
with machine learning techniques.
Our goal is to understand whether enriching text with linguistic information can improve incoherence detection in automatically generated narratives, guiding further research in the field. To do so, we propose incorporating RST or POS information into the texts to enrich the classifier's inputs. 

Moreover, we evaluate, using a Brazilian disinformation dataset called Fake\-TrueBR~\cite{chavarro2023faketruebr}, the hypothesis that textual incoherence can serve as an indirect indicator of disinformation. The main contributions of this work are:

\begin{itemize}
    \item We propose a strategy to enrich texts with additional symbols derived from RST and POS information, and investigate whether it enhances model accuracy in detecting incoherence in \mbox{stories}.

    \item We show that coherence classification can serve as a \textit{proxy} for detecting disinformation in offline scenarios, providing evidence of the practical applicability of coherence assessment. Our zero-shot cross-language pipeline also demonstrates that the knowledge can be reasonably transferred without further training on the new \mbox{language}.
\end{itemize}

\begin{figure*}[b]
    \centering
    \includegraphics[width=0.975\linewidth]{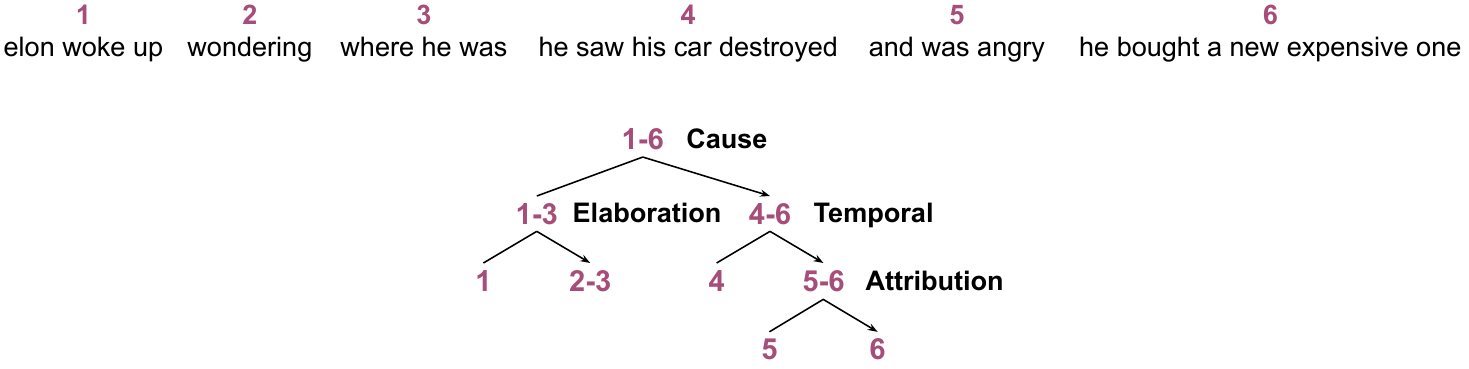}
    \caption{Example of RST relations represented as a tree. The EDUs are shown in the first row. Each non-leaf node represents a relation, where the nucleus is the child node given by the tail of the arrow (left) and the satellite is the arrowhead (right). For instance, there is a Cause relation between EDUs 1--3 (nucleus) and EDUs 4--6 (satellite).}
    \label{fig:ex_rst}
\end{figure*}

\begin{figure*}[t]
    \centering
    \includegraphics[width=.6\linewidth,clip,trim={0 0.75cm 0 0}]{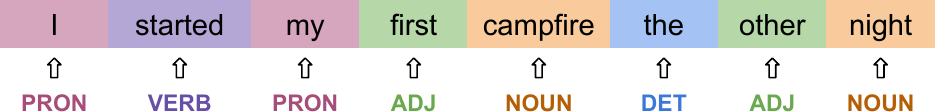}
    \caption{Example of POS tags.}
    \label{fig:ex_pos}
\end{figure*}

\section{Background and Related Work}
\label{sec:related}

In this section, we first revisit the RST, which provides the linguistic foundation for representing rhetorical relations in text. Then, we discuss POS tagging, which involves assigning grammatical class labels to each token in a text, providing syntactic information that complements the rhetorical structures considered in our approach. Finally, we review studies on coherence classification, focusing on approaches that integrate linguistic knowledge into language models to improve their ability to detect textual~coherence.

\paragraph{\textbf{Rhetorical Structure Theory}}(RST)~\cite{mann1988rhetorical} posits that, in addition to the explicit propositional content contained in the clauses of a text, there are implicit propositions, called \textit{relational propositions}, which emerge from the relations established between portions of the discourse. Skoufaki~\cite{skoufaki2020rhetorical} demonstrates that it is possible to identify coherence breaks when a text is analyzed according to the definitions and diagrams of RST. Relational propositions result from combinations of clauses or text segments that, beyond their explicit propositional meaning, convey implicit content responsible for emphasizing or constraining certain aspects within a sentence~\cite{mann1988rhetorical}. Thus, these propositions are linked not only to text organization but also to relationships within its thematic structure, suggesting that applying this theory to text generation could guide overall organization and the selection of subsequent sentences.

In RST, a relation identifies the connection between two or more non-overlap\-ping portions of text (Figure~\ref{fig:ex_rst}), called \textit{Elementary Discourse Units} (EDUs), which can be classified as nuclei or satellites. Within these relations, nuclei represent the most salient part or the essential information of the relation, while satellites provide supporting or background information.

\paragraph{\textbf{Part-Of-Speech (POS) Tagging}}~involves assigning grammatical class labels to each token in a text (Figure~\ref{fig:ex_pos}). These labels identify the syntactic function of each word in a sentence, enabling a deeper analysis of the grammatical structure and meaning of the text~\cite{pradhan2005semantic}. POS tagging is often performed using supervised learning models, in which each word is assigned to its grammatical class. These models learn linguistic patterns from training data and apply them to predict POS labels for new unannotated texts.

\paragraph{\textbf{Coherence Classification.}}~The literature identifies two main categories for incorporating linguistic knowledge into a classifier: (i) embedding knowledge through additional layers or architectural modifications; and (ii) processing the input/output to include linguistic information for the generation and classification model.\label{sec:backgroundclassf}

Considering the first category, Ke et al.~\cite{ke2019sentilare} propose combining POS information with the RoBERTa model by adding embedding layers that encode this information for sentiment analysis. Linguistic embeddings condition text processing rather than being directly integrated into the input.

Abhishek \cite{abhishek2021transformer} proposes four coherence-related tasks, including binary classification. Among the proposed models, the Vanilla Transformer relies on no extra linguistic information, and the Fact-aware Transformer combines text embeddings with linguistic ones (subject, verb, and object information) by adding an extra encoding layer. Both models are based on Longformer. In their experiments on the Grammarly Corpus of Discourse Coherence (GCDC) corpus \cite{lai2018discourse}, the Vanilla model outperforms the Fact-aware Transformer. Similarly, Liu and Strube~\cite{liu2025discourse} propose a fusion Transformer that combines text and linguistic embeddings for coherence classification. However, linguistic information comes from sentence-level relations given by the Penn Discourse TreeBank (PDTB). Unlike PDTB, RST enforces a hierarchical organization of relations. The authors evaluate their RoBERTa-based classifier with and without linguistic information, and find that, contrary to Abhishek~\cite{abhishek2021transformer}, linguistic information improves GCDC performance. \mbox{Liu et al.~\cite{liu2023modeling}} use linguistic information (relations between nouns) to build sentence graphs that feed a Graph Convolutional Network to assess textual coherence.

\begin{figure*}[t]
    \centering
    \includegraphics[width=0.95\linewidth]{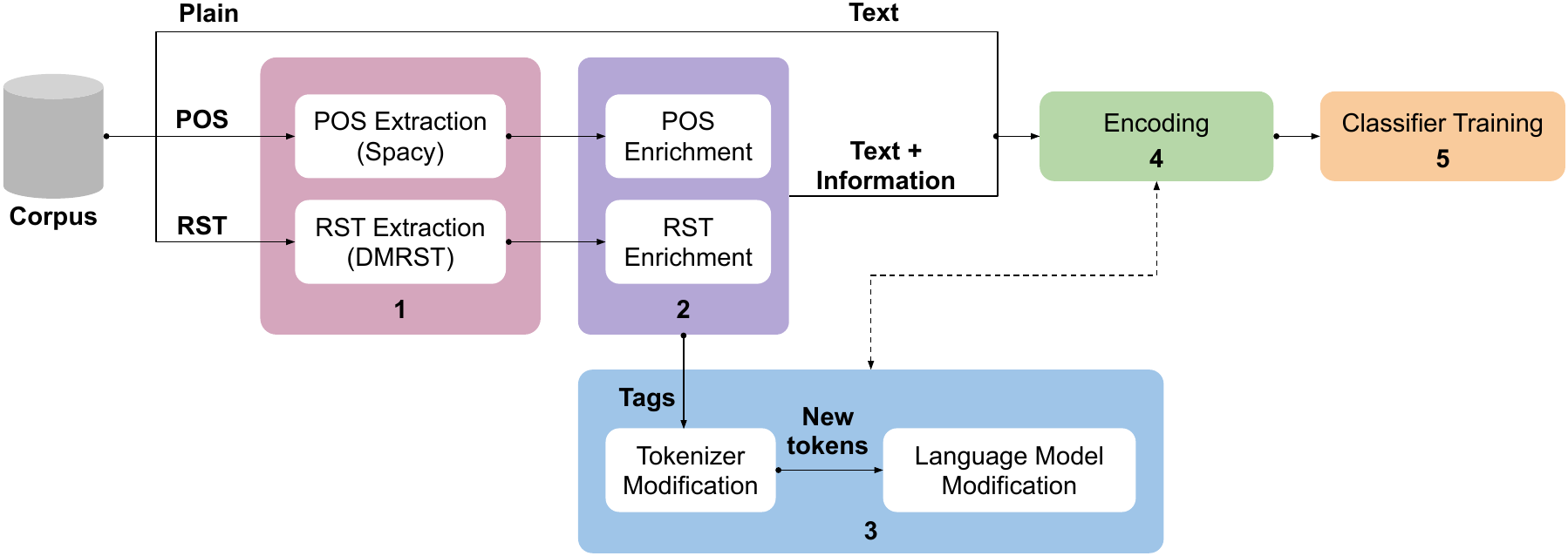}
    \caption{Overview of the training pipeline of our coherence classifier in three distinct versions: Plain, POS, and RST. The first follows a two-step training process, while the last two use a five-step pipeline that integrates syntactic/rhetorical information.}
    \label{fig:coherence_classifier}
\end{figure*}

The second category involves processing text to include linguistic information for the generation and classification models. Barzilay and Lapata \cite{barzilay2005modeling} propose a coherence assessment approach by augmenting texts with syntactic annotations indicating whether nouns (entities) are subjects, objects, neither, or absent in a sentence. They use this information to construct grids. The authors argue that some grid patterns are more common in coherent texts and encode these patterns into vectors to feed a machine learning method.  
They employ a Support Vector Machine method for coherence assessment. 
Chernyavskiy \cite{chernyavskiy2021correcting} leverages RST relations as markers during GPT-2 fine-tuning to improve text generation. For that, they integrate special tokens <R> into raw texts, where R indicates an RST's relation. Note that this requires only a minimal adjustment to the generative model to accept the special tokens, without any architectural modifications.

Our proposed method falls into the second category. To the best of our knowledge, our work is the first to assess the impact of augmented inputs in Transformer-based models for coherence classification. In the first category, there is evidence both supporting the effectiveness of linguistic information~\cite{liu2025discourse} and suggesting that it brings no improvement~\cite{abhishek2021transformer}. Therefore, we aim to determine which of these two holds in the second~\mbox{category.}

\section{Coherence Classifier}
\label{sec:method}

Our proposed methodology, illustrated in Figure~\ref{fig:coherence_classifier}, consists of five steps: (1)~extraction of POS/RST information; (2) text enrichment with POS/RST information; (3) tokenizer and language model modification; (4) text encoding; and (5) coherence classifier training. Steps~(4) and (5) consist of the usual text classification steps. The pipeline version we call Plain includes only these two steps. Text encoding (4) is based on the XLM-RoBERTa Longformer model, including both the tokenizer and the text embedding generator. For (5), we add an extra layer to the XLM-RoBERTa Longformer model to classify the text as coherent or incoherent. We detail steps (1), (2), and (3), proposed in this work, in the following \mbox{subsections.}

\subsection{Extraction of Linguistic Information}
\begin{figure*}[t]
    \centering
    \includegraphics[width=0.9\linewidth]{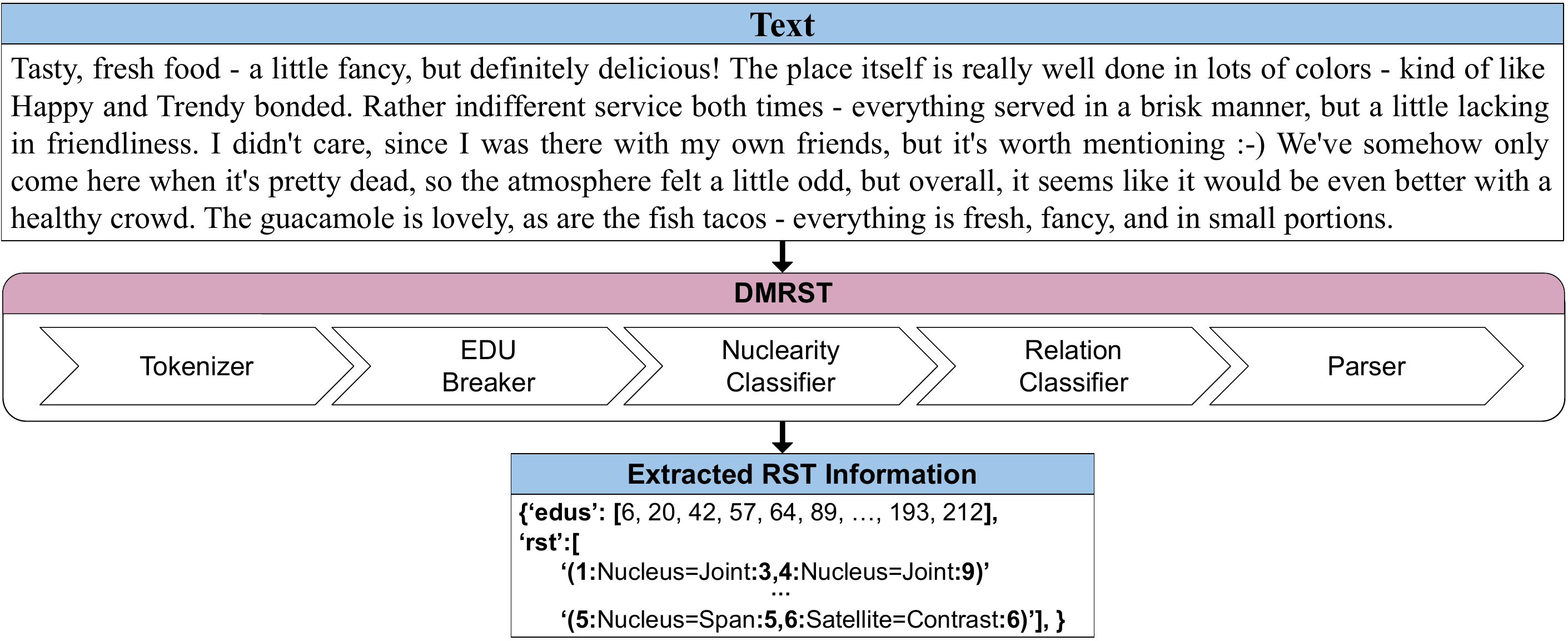}
    \caption{Example of RST extraction using the DMRST parser.}
    \label{fig:rstextraction}
\end{figure*}

\paragraph{\textbf{Extraction of RST Information.}}~We employ the DMRST parser~\cite{liu2021dmrst} to extract RST information. DMRST receives a text, separates it into tokens, and returns the EDU breaks and RST relations with nuclearity, requiring additional processing to obtain the EDUs. Figure~\ref{fig:rstextraction} illustrates the extraction process.

\begin{figure*}[ht]
    \centering
    \includegraphics[width=\linewidth]{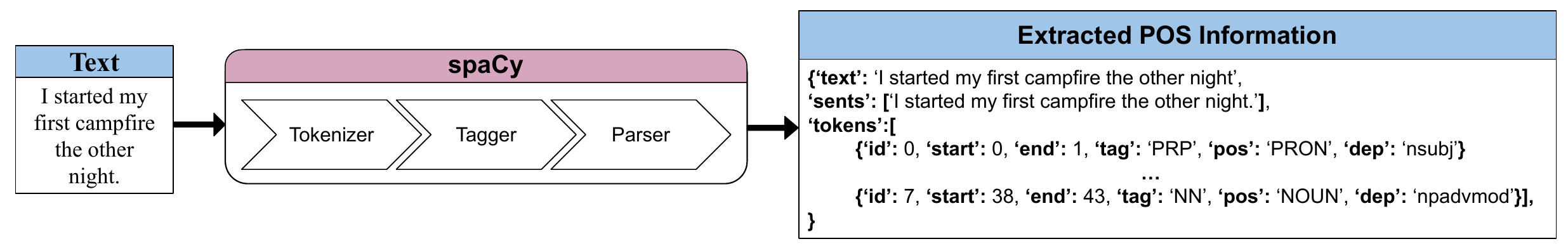}
    \caption{Example of POS extraction using the spaCy library.}
    \label{fig:posextraction}
\end{figure*}

\paragraph{\textbf{Extraction of POS Information.}}~To extract POS information, we use the spaCy parser~\cite{honnibal_spaCy_Industrial-strength_Natural_2020}. This Python library provides high-accuracy and efficient POS tagging models. We used the parser to annotate the words in the texts with their respective grammatical classes, which were used for the POS enrichment stage. Figure~\ref{fig:posextraction} shows an example of this process.

\subsection{Application of Text Enrichment}

Although both RST and POS text enrichment consist of adding preprocessed information to plain text, they differ in how this information is added.

\paragraph{\textbf{RST Enrichment.}}
Enriching texts with RST consists of preprocessing texts and adding new symbols to language models. For this purpose, we create special tokens that represent RST relations, nuclearity, and EDU breaks. Each token consists of nuclearity and relation, separated by a colon~(:), and enclosed by greater-than and less-than symbols (<>). The creation of these tokens and their positioning were inspired by the addition of special tokens to the Phi-3 model~\cite{abdin2024phi} to handle text excerpts that had some predefined formatting or structure. Since the RST parser has 19 relation categories and 2 nuclearity categories, 38 special tokens would be generated. However, in multinuclear relations (Joint, Same Unit, and Textual Organization), the nuclearity is the same for all participants considered as the nucleus. Thus, we generated only 35 tokens to represent all EDU separations, their relations, and their nuclearities. Table~\ref{tab:RSTMixtokens} shows a complete list of tokens. To represent EDU divisions, we add special tokens at the beginning and end of each EDU, depending on the RST relation. Although the order in which special tokens are added can influence the classification, we use the order returned by the parser, since there is no standard order. Table~\ref{tab:rstmixexample} shows the process of adding these~tokens.

\begin{table*}[t]
\setlength{\tabcolsep}{3pt}
\centering
    \caption{Tokens generated for the RST annotation: one nucleus (N) and one satellite (S) token per mononuclear relation.\vspace{-0.15cm}}
    \footnotesize
    \begin{tabular}{ll | ll}
    \hline
        \textbf{Relation} & \textbf{Token} & \textbf{Relation} & \textbf{Token}\\
    \hline
        Topic Change & <N/S:Topic-Change> & Attribution & <N/S:Attribution>\\
        Background & <N/S:Background> & Evaluation & <N/S:Evaluation>\\
        Contrast & <N/S:Contrast> & Condition & <N/S:Condition>\\
        Summary & <N/S:Summary> & 
        Explanation & <N/S:Explanation>  \\ 
        Comparison & <N/S:Comparison> & Elaboration & <N/S:Elaboration>\\
        Temporal & <N/S:Temporal> & Span & <N/S:Span>\\
        Enablement & <N/S:Enablement> & Same Unit & <N:Same-Unit>\\
        Cause & <N/S:Cause> & Joint & <N:Joint>\\
        Topic Comment & <N/S:Topic-Comment> & Textual & <N:TextualOrganization>\\
        Manner Means & <N/S:Manner-Means> & Organization\\
    \hline
    \end{tabular}
    \label{tab:RSTMixtokens}
\end{table*}

\begin{table*}[t]
\setlength{\tabcolsep}{4pt}
    \caption{Example of plain text from which RST was extracted and the annotated text after the enrichment.\vspace{-0.15cm}}
    \centering
    \begin{tabular}{>{\raggedright}p{0.28\linewidth}|>{\raggedright\arraybackslash}p{0.675\linewidth}}
    \hline
    \textbf{Plain Text}     & \textbf{RST Enrichment}  \\
    \hline
I chose to stay here because it’s so close to the convention center for my son’s bball tourny. The check in process was sooo slow. Even if I requested for a double bedroom a while back it was not given to us and then we have to change rooms the next day and the room they gave us smells so bad of cigarette smoke and it’s supposed to be a non-smoking room! & 
<N:span><N:span>I chose to stay here<N:span><S:Explanation>because its so
close to the convention center for my son’s bball tourny. <N:span><S:Explanation><N:Joint>The check in process was sooo slow.<N:Joint><S:Contrast><N:Joint><N:Joint>Even if I requested for a double bedroom a while back<S:Contrast><N:Span>it was not given to us<N:Span><N:Joint><N:Joint><N:Joint>and then we have to change rooms the next day<N:Joint><N:Span><N:Same-Unit><N:Joint><N:Joint>and the room<N:Span><S:Elaboration>they gave us<N:Same-Unit><S:Elaboration><N:Same-Unit>smells so bad of cigarette smoke<N:Joint><N:Same-Unit><N:Joint>and it’s supposed to be a non-smoking room!<N:Joint><N:Joint><N:Joint><N:Joint>\\
\hline
    \end{tabular}
    \label{tab:rstmixexample}
\end{table*}

\paragraph{\textbf{POS Enrichment.}}~Enriching texts with POS is similar to enriching with RST, but the tokens only represent the grammatical classes of the words. We drew inspiration from the preprocessing presented by~\mbox{Wang et al.~\cite{wang2024can}}, which merges this information with the text, and adapted it to our context. Each token is formed by an underscore (\_) followed by one of the 17 grammatical classes returned by the POS parser. The parts of speech used by the parser are the same as those defined in the Universal POS tags (UPOS) for treebank annotations in a multilingual setting~\cite{petrov2012universal}. Table~\ref{tab:postokenslist} shows the complete list of tokens. The tokens are added after each word of the original text, with no space between them. Table~\ref{tab:posmixexample} shows an example of adding special tokens.

\begin{table*}[t]
\setlength{\tabcolsep}{4pt}
    \centering
    \caption{List of POS symbols generated by the spaCy library, their meaning, and the generated token.\vspace{-0.15cm}}
    \begin{tabular}{lll|lll}
    \hline
      \textbf{UPOS}   &  \textbf{Meaning} & \textbf{Token} & \textbf{UPOS}   &  \textbf{Meaning} & \textbf{Token}\\
      \hline
       SCONJ  & Subordinate & \_SCONJ & CCONJ & Coordinating  & \_CCONJ\\
       & conjunction & & & conjunction\\
       NUM & Numeral & \_NUM & PROPN & Proper noun & \_PROPN\\
       X & Others & \_X & ADV & Adverb & \_ADV\\
       NOUN & Noun & \_NOUN & ADJ & Adjective & \_ADJ\\
       PUNCT & Punctuation & \_PUNCT & PRON & Pronoun & \_PRON\\
       AUX & Auxilliar & \_AUX & SYM & Symbol & \_SYM\\
       ADP & Adposition & \_ADP & PART & Particle & \_PART\\
       DET & Determinant & \_DET & INTJ & Interjection & \_INTJ\\
       VERB & Verb & \_VERB & SPACE & Space between words & None\\
       \hline
    \end{tabular}
    \label{tab:postokenslist}
\end{table*}

\begin{table*}[ht]
    \centering
\setlength{\tabcolsep}{3pt}
    \caption{Example of plain text from which POS was extracted and the annotated text after the enrichment.\vspace{-0.15cm}}
    \begin{tabular}{>
    {\raggedright}p{0.27\linewidth}|>{\raggedright\arraybackslash}p{0.69\linewidth}}
    \hline
    \textbf{Plain Text}     & \textbf{POS Enrichment}  \\
    \hline
I chose to stay here because it’s so close
to the convention center for my son’s
bball tourny. The check in process was
sooo slow. Even if I requested for a
double bedroom a while back it was not
given to us and then we have to change
rooms the next day and the room they
gave us smells so bad of cigarette smoke
and it’s supposed to be a non-smoking
room!  & 
I\_PRON chose\_VERB to\_PART stay\_VERB here\_ADV because\_SCONJ it\_PRON ’s\_AUX so\_ADV close\_ADJ to\_ADP
the\_DET convention\_NOUN center\_NOUN for\_ADP my\_PRON son\_NOUN ’s\_PART bball\_NOUN tourny\_NOUN
.\_PUNCT The\_DET check\_NOUN in\_ADP process\_NOUN was\_AUX sooo\_ADV slow\_ADJ .\_PUNCT Even\_ADV
if\_SCONJ I\_PRON requested\_VERB for\_ADP a\_DET double\_ADJ bedroom\_NOUN a\_DET while\_NOUN back\_ADV
it\_PRON was\_AUX not\_PART given\_VERB to\_ADP us\_PRON and\_CCONJ then\_ADV we\_PRON have\_VERB
to\_PART change\_VERB rooms\_NOUN the\_DET next\_ADJ day\_NOUN and\_CCONJ the\_DET room\_NOUN
they\_PRON gave\_VERB us\_PRON smells\_VERB so\_ADV bad\_ADJ of\_ADP cigarette\_NOUN smoke\_NOUN
and\_CCONJ it\_PRON ’s\_AUX supposed\_VERB to\_PART be\_AUX a\_DET non\_ADJ -\_ADJ smoking\_ADJ room\_NOUN
!\_PUNCT\\
    \hline
    \end{tabular}
    \label{tab:posmixexample}
\end{table*}

\subsection{Changing Tokenizer and Model}

\paragraph{\textbf{Tokenizer.}} To support the special tokens for the enrichment methods, new tokens must be added to the tokenizer's vocabulary. Treating additional tokens as special allows them to be ignored during decoding and not split into subwords, while aiding understanding of the model's results. Not splitting them into subwords is particularly important for the proposed methods, since splitting these tokens can alter their meaning and impair classification of long texts, as we may exceed the model's limit by adding more tokens.

\paragraph{\textbf{Model.}} The tokenizer update leads to a model update, since the model's token embedding matrix must contain the same number of rows as the tokenizer's vocabulary. The vectors added to the end of the embedding matrix are initialized randomly, impairing the model's convergence. To alleviate this problem and allow the model to learn the vectors of the additional tokens more quickly, the weights of these vectors were replaced with values corresponding to the average of the weights of other vectors in the embedding matrix, a common practice when extending the \mbox{vocabulary~\cite{hewitt2021initializing}.}

\paragraph{\textbf{Loss Function.}}~We used the weighted binary cross-entropy,  an extension of binary cross-entropy that assigns different weights to each class, so that the minority class has a greater weight than the majority class (Equation~\ref{eq:weighted_binary_cross_entropy}). We chose this loss function because the GCDC corpus is imbalanced, with the coherent class as the majority and the incoherent class as the \mbox{minority.} 
\begin{equation} \label{eq:weighted_binary_cross_entropy}
\small
    W = -\frac{1}{N}\sum^N_{i=1}\left[w_1y_i\log(p_i)+w_0(1-y_i)\log(1-p_i)\right],
\end{equation}
\noindent where $w_1$ refers to the weight for the positive class (\textit{i.e.}, incoherent texts), $w_0$ is the weight for the negative class (\textit{i.e.}, coherent texts), $y_i\in\{ 0,1\}$ is the binary label of example $i$, $p_i$ is the probability predicted by the model for the positive class, and $N$ is the total number of examples.
\section{Experiments}
\label{sec:experiments}

\subsection{Datasets}

\paragraph{\textbf{Grammarly Corpus of Discourse Coherence (GCDC).}} The GCDC corpus~\cite{lai2018discourse} contains English texts from four sources: Yahoo online forum posts, Yelp business reviews, emails from Hillary Clinton’s office, and emails sent by Enron employees. Three specialists with previous annotation experience annotated each text. 
Each annotator classified the texts into three coherence levels: low, medium, or high. The coherence label was assigned to each text according to the most frequent label among the expert annotators. In case of disagreement among the annotators, the text was labeled with medium coherence.

We changed the corpus to a binary version, in which texts with high coherence were deemed coherent and those with low coherence were deemed incoherent. In our view, the presence of even one type of incoherence is enough to classify a text as incoherent, there being no middle ground for what might be considered medium coherence.

\paragraph{\textbf{FakeTrueBR.}} The FakeTrueBR corpus~\cite{chavarro2023faketruebr} contains about 3500 Brazilian Portuguese texts from news articles, social media posts, and other formats of digital media, collected via a content crawler. The corpus is designed for disinformation and fact-checking analysis and stands out for pairing texts that show correspondence between disinformation and real news. We use it exclusively for zero-shot evaluation, without training or validation.

We evaluate our method on news texts (Brazilian Portuguese) based on our hypothesis that coherence-trained models can detect disinformation, given the presence of contradictions and informal language in these texts and in texts deemed incoherent.

\subsection{Validation and Evaluation}

We evaluate using balanced accuracy with mean and standard deviation across five runs, complemented by the Brier Score Loss~\cite{brier1950verification}, which assesses how well a model's predicted probabilities align with the observed outcomes. We report the results by corpus and its \mbox{subsets.}
\section{Results}
\label{sec:results}

We used the XLM-RoBERTa model to train the coherence classifier on the GCDC corpus. We repeated each of the three proposed pipelines (Section~\ref{sec:method}) five times. We also conducted experiments with different numbers of scheduler cycles and different learning rates, but only for the Plain pipeline. We replicated the best values to the other pipelines to reduce the experiment load. We employed the best model on the English corpus on the Brazilian corpus.

\subsection{GCDC}

Table~\ref{tab:GCDCresults} presents balanced accuracy results for GCDC. First, we varied the number of cycles for learning rate (LR) restarts, keeping the LR at $5\times10^{-5}$. With 10~cycles, Plain exceeded $70.0\%$ balanced accuracy, surpassing Abhishek's~\cite{abhishek2021transformer} accuracy result, which tends to be higher than the balanced. We progressively reduced the number of cycles until reaching 1 to observe the classifier's behavior. Lastly, we reduced the LR as recommended by Godbole \cite{tuningplaybookgithub}. This last option had the highest average balanced accuracy across all pipelines. The Plain pipeline consistently outperformed the RST and POS, with RST ranking second.

\begin{table*}[t]
\setlength{\tabcolsep}{2pt}
    \centering
    \caption{Main results for the experiments performed on the GCDC corpus. Highlighted values represent the best result obtained, and underlined values represent the best score per pipeline. Cyc.: Cycles; LR: Learning Rate; Acc.: Accuracy.}
    \begin{tabular}{c c c c c c c c c c c}
    \hline
     \multicolumn{3}{c}{Plain} & & \multicolumn{3}{c}{RST} & & \multicolumn{3}{c}{POS}\\
     \textbf{Cyc.} & \textbf{LR} & \textbf{Acc. (\%)} & & \textbf{Cyc.} & \textbf{LR} & \textbf{Acc. (\%)} & & \textbf{Cyc.} & \textbf{LR} & \textbf{Acc. (\%)}\\
    \hline
         $10$ & $5\times10^{-5}$ & $71.2$ \scriptsize{$\pm~2.7$} & & $2$ & $5\times10^{-5}$ & $67.1$ \scriptsize{$\pm~0.3$} & & 1 & $1\times10^{-5}$ & $\underline{68.1}$ \scriptsize{$\pm~0.4$}\\
         $8$ & $5\times10^{-5}$ & $72.0$ \scriptsize{$\pm~0.2$} & & $1$ & $5\times10^{-5}$ & $67.2$ \scriptsize{$\pm~0.2$}\\
         $5$ & $5\times10^{-5}$ & $71.7$ \scriptsize{$\pm~0.2$} & & $1$ & $1\times10^{-5}$ & $\underline{69.0}$ \scriptsize{$\pm~0.2$}\\
         $3$ & $5\times10^{-5}$ & $71.5$ \scriptsize{$\pm~0.3$}\\
         $2$ & $5\times10^{-5}$ & $71.7$ \scriptsize{$\pm~0.2$}\\
         $1$ & $5\times10^{-5}$ & $71.1$ \scriptsize{$\pm~0.4$}\\
         $1$ & $1\times10^{-5}$ & $\underline{\mathbf{72.4}}$\scriptsize{$\pm~0.1$}\\     
    \hline
    \end{tabular}
    \label{tab:GCDCresults}
\end{table*}

\begin{table*}[t]
\begin{minipage}[t]{0.48\textwidth}
\setlength{\tabcolsep}{6pt}
    \centering
    \caption{Balanced accuracy for each pipeline with respect to each subset of the GCDC corpus. Highlighted values represent the best result, and underlined values represent the best score per pipeline.}
    \begin{tabular}{l c c c}
    \hline
        \multirow{2}{*}{\textbf{Subset}} & \multicolumn{3}{c}{\textbf{Accuracy (\%)}} \\
         & Plain & RST & POS\\
        \hline
        Clinton & $73.1$ \scriptsize{$\pm~0.2$} & $64.6$ \scriptsize{$\pm~0.5$} & $67.7$ \scriptsize{$\pm~0.2$}\\
        Yelp & $64.5$ \scriptsize{$\pm~0.4$} & $59.8$ \scriptsize{$\pm~0.3$} & $59.1$ \scriptsize{$\pm~0.1$}\\
        Enron & $72.4$ \scriptsize{$\pm~0.2$} & $\underline{70.6}$ \scriptsize{$\pm~0.2$} & $70.4$ \scriptsize{$\pm~0.1$}\\
        Yahoo & $\underline{\mathbf{74.1}}$\scriptsize{$\pm~0.2$} & $70.5$ \scriptsize{$\pm~0.5$} & $\underline{72.7}$ \scriptsize{$\pm~0.3$}\\
         \hline
    \end{tabular}
    \label{tab:GCDCresultssub}
\end{minipage}
\hfill
\begin{minipage}[t]{0.48\textwidth}
\setlength{\tabcolsep}{6pt}
    \centering
     \caption{Brier Score Loss for each subset of the GCDC corpus. The metric was calculated separately for each subset and globally, \textit{i.e.}, using the entire corpus. Highlighted values represent the best result, and underlined values represent the best score per pipeline.}
    \begin{tabular}{l c c c}
    \hline
        \multirow{2}{*}{\textbf{Subset}} & \multicolumn{3}{c}{\textbf{Brier Score Loss $\downarrow$}} \\
         & Plain & RST & POS\\
        \hline
        Clinton & $0.300$ & $0.272$ & $0.236$\\
        Yelp & $\underline{0.296}$ & $0.268$ & $\underline{\mathbf{0.234}}$\\
        Enron & $0.303$ & $\underline{0.263}$ & $0.239$\\
        Yahoo & $0.422$ & $0.368$ & $0.260$\\
        All & $0.331$ & $0.294$ & $0.243$\\
         \hline
    \end{tabular}
    \label{tab:brierscoreloss}
\end{minipage}
\end{table*}

We also analyze results by subset (Table~\ref{tab:GCDCresultssub}) and hypothesize that the model performs differently across text types. Yelp consistently shows the worst performance, likely because its annotations tend to label colloquial writing as incoherent and norm-following text as coherent. Per-subset results also show that, although the RST pipeline outperforms the POS one overall, this does not hold for specific styles such as Clinton and Yahoo. 

\begin{figure*}[t]
    \centering
    \includegraphics[width=\linewidth,clip,trim={0 1.3cm 0 0 }]{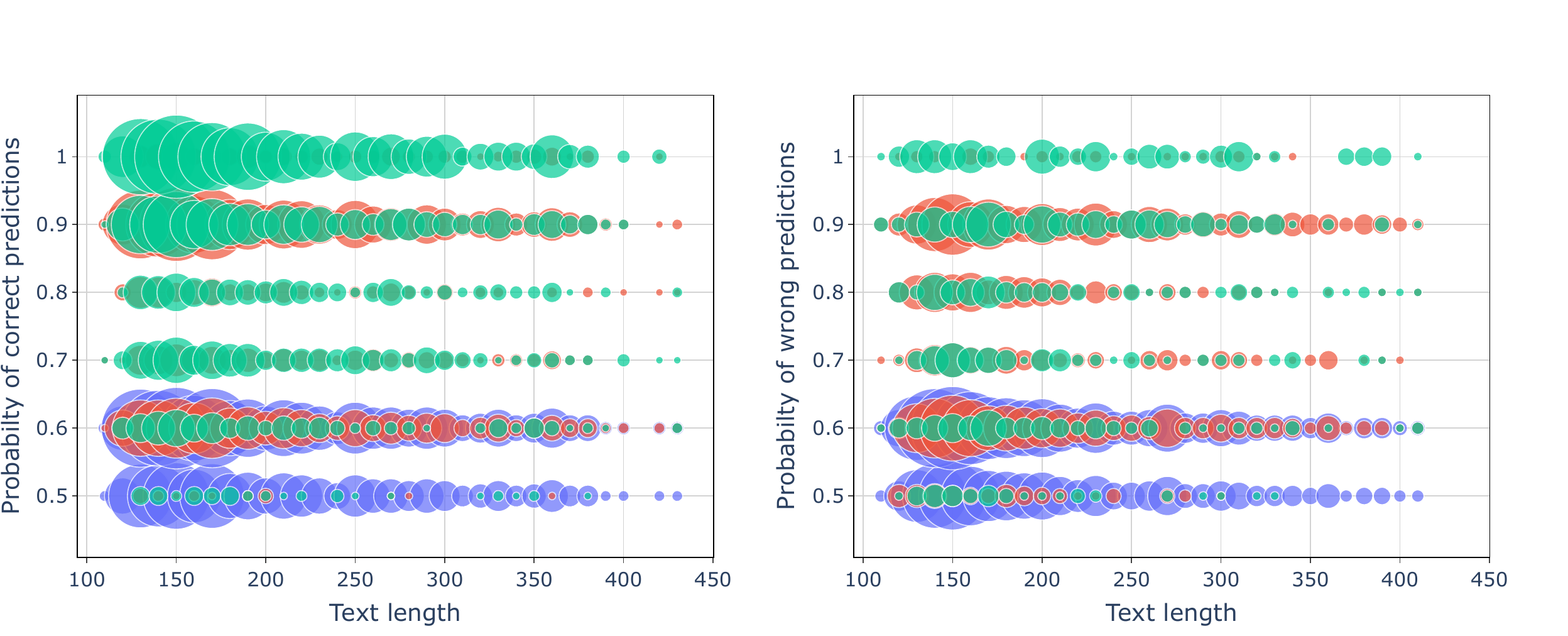}
    \caption{Probability distribution of correct and incorrect predictions on the GCDC corpus with Vanilla~\textcolor[RGB]{82,190,128}{\textbullet}, RST~\textcolor[RGB]{210,100,75}{\textbullet}, and POS~\textcolor[RGB]{150,140,210}{\textbullet} pipelines. The circles' diameters are proportional to the number of texts corresponding to each length-probability~pair.}
    \label{fig:prob_dist}
\end{figure*}

Table \ref{tab:brierscoreloss} presents the results for the Brier Score Loss. For this metric, we conducted experiments for each subset and for the entire corpus. This score ranges from 0 (best possible) to 1 (worst case). It indicates the model's confidence level when making correct or incorrect predictions. Unlike balanced accuracy, the POS pipeline performed best overall, while the Plain pipeline performed worst. Therefore, although the Plain pipeline achieved the best accuracy, it presented higher confidence levels even for incorrect predictions. Figure~\ref{fig:prob_dist} illustrates this behavior.

\subsection{FakeTrueBR}

For the FakeTrueBR corpus, we carried out only zero-shot experiments, using models trained on the GCDC. To compare results, we calculated the balanced accuracy (Table~\ref{tab:FakeTrueresults}).  The Plain pipeline performed best when trained on the GCDC corpus, followed by the RST pipeline. Although far from the average accuracy reported in the original paper ($94.5\%$)~\cite{chavarro2023faketruebr}, we achieved around 3/4 of the texts right in our best case while training models for other tasks and in \mbox{other languages.}

\begin{table*}[t]
\setlength{\tabcolsep}{6pt}
    \centering
    \caption{Obtained results with the balanced accuracy metric for the FakeTrueBR corpus in a zero-shot scenario. Highlighted values represent the best result, and underlined values represent the best score per pipeline.\vspace{-0.15cm}}
    \begin{tabular}{c c c}
    \hline
    \multicolumn{3}{c}{\textbf{Accuracy (\%)}} \\
     Plain & RST & POS\\
    \hline
    $\textbf{73.1}$\scriptsize{$\pm~0.8$} & $\underline{61.5}$ \scriptsize{$\pm~0.5$} & $59.5$ \scriptsize{$\pm~0.6$}\\
    \hline
    \end{tabular}
    \label{tab:FakeTrueresults}
\end{table*}
\section{Discussion}
\label{sec:disc}

\paragraph{\textbf{GCDC.}}~It may seem unexpected that enriched text underperforms plain text. In this subsection, we offer some justifications for this performance. 

For POS enrichment, we understand that the type of information added is not well aligned with the objective of coherence evaluation, considering that incoherent texts can be grammatically correct, \textit{i.e.}, the relationship between coherence and standard writing norms is not one of strict inclusion \cite{tanya-conditions}. This may lead the model to generate false positives and false negatives, which explains an accuracy below that of the Plain pipeline.

On the other hand, Table~\ref{tab:GCDCresultssub} shows that for the Yahoo and Clinton subsets of the GCDC, the POS enrichment method performs better than the RST enrichment, which one would expect to be incorporating more relevant information for evaluating textual coherence. We understand that this is not due to a performance gain from the POS pipeline, but to poor performance from the RST one, due to structural incompatibility between the RST and the Transformer architecture. 

From a mathematical perspective on the Transformer architecture, we argue that it is a non-isomorphic operator in the $\mathbb{R}^{d\times n}$ Euclidean space of sequences of $n$ vectors of length $d$. Indeed, an isomorphism is a bijective operator. It has been shown that decoder only Transformers are injective, except on Lebesgue measure-zero subsets, when taken as discrete-to-continuous maps from prompts~$s$ in a finite vocabulary $\mathcal{V}\subset \mathbb{R}^{d\times n}$ to hidden states in $\mathbb{R}^{d\times n}$~\cite{nikolaou2026injective}. In such a point-of-view, the Transformer cannot be surjective, as a map from a countable subset of $\mathbb{R}^{d\times n}$ cannot map to all uncountable elements of $\mathbb{R}^{d\times n}$ \cite{rudin1976principles}. See the \nameref{appendix} for more detail.

When visualizing the RST structure as a directed acyclic graph, the lack of isomorphism implies that the Transformer does not preserve the graph's structure and relations. In this sense, the RST structure is flattened when passed through the Transformer, and the added information is retrieved as noise when training, thus the underperformance in comparison to the Plain pipeline.

\paragraph{\textbf{FakeTrueBR.}}~For the zero-shot experiments on the FakeTrueBR dataset, we verified that the Plain pipeline also performed better, achieving an accuracy comparable to its coherence assessment. In the context of domain transfer, we expected that the enriched pipelines would underperform the Plain one, as the latter performs better at assessing coherence. On the other hand, RST and POS pipelines performed even worse than their coherence counterparts, with POS enrichment being the worst of the two.

For POS enrichment, we understand that, as misleading and factually inaccurate texts are often syntactically well written to divert readers from their inaccuracy, the pipeline often predicts disinformation as real news. For RST enrichment, the performance is expected when taking it as noise for the model.  
 
Nevertheless, the experiments with the FakeTrueBR corpus shows an underlying connection between textual incoherence and misleading content, as the Plain pipeline reached an accuracy of $3/4$ of the dataset, despite being fine-tuned on another task and in another language.
\section{Conclusion and Future Work}
\label{sec:conclusion}

This paper investigated textual coherence classification in narratives by integrating RST and syntactic (POS) information into language models. Our proposed methodology combines linguistic theories with machine learning techniques. We showed that the plain-text model achieves the best accuracy compared with RST- and POS-enriched texts, and we provided mathematical and linguistic reasons for this. We also showed that expanding the model's vocabulary with RST- and POS-derived symbols improved incoherence prediction confidence in some scenarios.

In this sense, we conclude that simply enriching texts with linguistic structure does not serve well the purpose of assessing coherence (at least not with RST and POS). Therefore, for future work, we intend to focus on incorporating linguistic features via additional embedding layers or other networks, such as Graph or Convolutional Neural Networks, which, as we discussed in Section~\ref{sec:backgroundclassf}, achieve superior results in text generation and classification. One way to advance in this direction would be to replicate these strategies for RST. We also hypothesize that enriching text with RST would yield better results when coupled with architectural modifications that account for hierarchical structures, such as Poincaré Embeddings~\cite{nickel-poincare-embeddings} and Graph-aware Isomorphic Attention~\cite{buehler-graph-aware}.

Recent works \cite{liu2025discourse,liu-strube-2025-joint} also address the use of the PDTB~\cite{prasad-etal-2008-penn,webber-2019-pdtb}, a linear discourse annotation framework, and are therefore structurally aligned with the Transformer architecture. Thus, enriching text with PDTB, instead of RST or POS, could achieve better results. We intend to implement this enrichment and repeat the experiments, hypothesizing that its performance is directly related to the overall adequacy of the text enrichment technique.

On the other hand, experiments showed that the models not only serve for incoherence detection but also presented promising results in zero-shot tests on FakeTrueBR. However, the correlation between coherence and disinformation, while statistically relevant, demands further investigation to distinguish between intentionally misleading texts and those that are simply poorly structured. For the future, it would be relevant to replicate the zero-shot experiments in the same language domain on which the classifier was~trained.
\section{Ethics Statements and Limitations}
\label{sec:ethics}

This study adheres to the ethical principles of Natural Language Processing research. The proposed models are intended solely for academic research, involve no human subject data, and use only publicly available datasets. The following limitations and statements~\mbox{apply}.

\paragraph{\textbf{Annotated Data.}} Although our models are multilingual, the DMRST parser \cite{liu2021dmrst}  supports only six languages: English, Portuguese, Spanish, German, Dutch, and Basque. This limitation arises not only from the parser itself but also from the scarcity of available annotated datasets. Annotating RST requires expert knowledge, and even trained annotators may produce inconsistent structures due to differences in interpretation or linguistic nuances. 

\paragraph{\textbf{Textual Coherence and Arguments.}} We define discourse as structured text composed of coherent sentences, where coherence (local, global, or topical) ensures semantic connectivity and non-contradiction.  However, coherent texts may still contain biased, misleading, or factually incorrect content. For instance, some GCDC corpus examples express xenophobic views that are structurally coherent but incoherent with our worldview.

\paragraph{\textbf{Carbon and Water Usage Estimator.}} This experiment consumed 385~kWh, with an estimated footprint of 37.86 kgCO$_2$e and 7,154.97 L of water in Brazil, as calculated with wAIter (beta version) \cite{breder2025waiter}. Since we used three GPUs, the total estimated consumption was 1,155.00 kWh, corresponding to a footprint of 113.58 kgCO$_2$e and 21,464.91 L of water.

\paragraph{\textbf{Generative AI and AI-assisted Technologies.}} The authors occasionally made use of DeepL Translate, ChatGPT and Gemini during the preparation of this paper to assist with language refinement and the editing process. The authors carefully reviewed and revised all text, assuming full responsibility for the final content.

\bibliographystyle{splncs04}
\bibliography{bibliography}

\subsection*{Appendix}
\label{appendix}
In a continuous theoretical setting, such as proposed in the original Transformers paper \cite{vaswani2017attention}, a decoder-only Transformer with $L$ layers is the composition of $L$ Transformer-blocks\cite{thickstun2021transformer}. A Transformer block is a parametrized function $f_\theta: \mathbb{R}^{d\times n} \to \mathbb{R}^{d\times n}$. For an input $x = (x_1, x_2, \dots, x_{d\times n}) \in \mathbb{R}^{d\times n}$, $f_\theta (x) = z$ is calculated as follows: the Query ($Q$), Key ($K$) and Value~($V$) matrices are trainable weights over the \mbox{input:}
\[
\resizebox{0.45\textwidth}{!}{%
    $Q^{(h)}(x_i) = W^T_{h,q} x_i, \  K^{(h)}(x_i) = W^T_{h,k} x_i, \  V^{(h)}(x_i) = W^T_{h,v} x_i,$
    }
\]
for each head $h$ and trainable weight matrices $W$. Then, the self-attention between vectors $x_i$ and $x_j$ is defined as
\[
\alpha_{i,j}^{(h)} = \text{softmax}\left(\frac{\langle Q^{(h)}\rangle (x_i), K^{(h)}(x_j)}{\sqrt{k}}\right).
\]

Then, the self-attention is summed over weights per head
\[
u_i' = \sum_{h=0}^H W_{c,h}^T\sum_{j=1}^n \alpha_{i,j}^{(h)}V^{(h)}(x_j),
\]
and normalized
\[
u_i = \text{LayerNorm}(x_i+u_i';\gamma_1,\beta_1).
\]
This vector goes through an activation function (such as ReLU):
\[
z_i'= W^T_2\text{ReLU}(W_1^Tu_i),
\]
which is again normalized to obtain the output of the Transformer block
\[
z_i = \text{LayerNorm}(u_i+z_i';\gamma_2, \beta_2).
\]
The LayerNorm function is defined as
\[
\begin{split}
\text{LayerNorm}(z; \gamma, \beta) &= \gamma\frac{(z-\mu_z)}{\sigma} + \beta,\\
\mu_z &= \frac{1}{k}\sum_{i=0}^k z_i,\\
\sigma &= {\sqrt{\frac{1}{k}\sum_{i=1}^k(z_i-\mu_z)^2}}.
\end{split}
\]
The Transformer is then the composition of $L$ Transformer blocks, $f_{\theta_L}\circ\dots\circ f_{\theta_1} \in \mathbb{R}^{d\times n}$.

In these settings, we aim to prove the following theorem:

\paragraph{\textbf{Theorem.}} Let $f_\theta: \mathbb{R}^{d\times n} \to \mathbb{R}^{d\times n}$ be the continuous decoder-only Transformer mapping sequences of $n$ continuous token embeddings of dimension $d$ to their final hidden states. Then, $f$ is not surjective.

\paragraph{\noindent\textbf{Proof.}} Our proof consists in decomposing the Transformer $f$ as $f = \phi \circ g$, where $g$ represents the composition of all architectural components before the final normalization function, represented by $\phi$. It suffices to prove that the outermost function $\phi$ is not surjective to prove that $f$ is not surjective.

For the LayerNorm, the normalized vector $z = \frac{x - \mu}{\sigma}$ is such that it lies on a hyperplane orthogonal to the identity vector, that is, $\sum_{i=1}^dz_i = 0$. The normalized vector also lies on a hypersphere of radius $\sqrt{d}$, that is, $||z||^2_2 =d$, where $||\cdot||_2$ is the euclidean norm. Then, the normalized vector lies on the manifold $M_i$, which is the intersection of a hyperplane with a hypersphere. Thus, the manifold where the normalized vector lies has dimension $d-2$ \cite{sommerville1929geometry}. 

Since the map $\phi$ applies individually to the entire sequence of $n$ vectors, the full image space manifold is the Cartesian product of each manifold $M_i$:
\[
M = M_1\times M_2 \times \dots \times M_n.
\]
As such, the dimension of the image is
\[
\text{dim}(M) = \sum_{i=0}^{n} \text{dim}(M_i).
\]
Since each $M_i$ has dimension $d-2$, $\text{dim}(M) = n(d-2) = nd-2n$. Because in any practical Transformer $n \geq 1$ and $d \geq 2$, we have
\[
\text{dim}(M) =nd-2n < nd = \text{dim}(\mathbb{R}^{d\times n}).
\]
Therefore, the image of $f$, denoted $\text{Im}(f) = \text{Im}(\phi\circ g) \subset \text{Im}(\phi)\subset M$, is constrained to a submanifold whose dimension is less than the dimension of the codomain. By properties of Lebesgue measure, any smooth manifold $M\subset \mathbb{R}^n$ of dimension $m<n$ has a Lebesgue measure of zero in $\mathbb{R^n}$ \cite{munkres2018analysis}. As such, $\text{Im}(f)$ has measure zero in $\mathbb{R}^{d\times n}$, implying that $\text{Im}(f)$ is a proper subset of $\mathbb{R}^{d\times n}$, and $f$ is not surjective. $\square$

On the other hand, when taking the Transformer as a discrete-to-continuous operator, Cantor's Theorem states that the operator cannot be surjective \cite{rudin1976principles}. An operator as such is the one presented in \cite{nikolaou2026injective}, defined by $f: \mathcal{V}^{\leq K} \times \mathbb{R}^p \to \Delta^{|\mathcal{V}|-1}$, that is, an operator from the finite vocabulary set with $p$ parameters to the simplex of dimension $|\mathcal{V}|-1$. Since the simplex is an uncountable subset of $\mathbb{R}^n$, the map $f$ cannot be surjective and, thus, it is not an isomorphism.
\end{multicols}
\end{document}